\let\TeXyear\year
\documentclass{ieeeaccess}
\let\Accessyear\year
\let\year\TeXyear
\definecolor{accessblue}{cmyk}{1,0.3,0,0.2}
\usepackage{cite}
\usepackage{amsmath,amssymb,amsfonts}
\usepackage{algorithmic}
\usepackage{graphicx}
\usepackage{textcomp}

\usepackage{bm}
\makeatletter
\AtBeginDocument{\DeclareMathVersion{bold}
\SetSymbolFont{operators}{bold}{T1}{times}{b}{n}
\SetSymbolFont{NewLetters}{bold}{T1}{times}{b}{it}
\SetMathAlphabet{\mathrm}{bold}{T1}{times}{b}{n}
\SetMathAlphabet{\mathit}{bold}{T1}{times}{b}{it}
\SetMathAlphabet{\mathbf}{bold}{T1}{times}{b}{n}
\SetMathAlphabet{\mathtt}{bold}{OT1}{pcr}{b}{n}
\SetSymbolFont{symbols}{bold}{OMS}{cmsy}{b}{n}
\renewcommand\boldmath{\@nomath\boldmath\mathversion{bold}}}
\makeatother

\def\BibTeX{{\rm B\kern-.05em{\sc i\kern-0.025em b}\kern-.08em
    T\kern-.1667em\lower.7ex\hbox{E}\kern-0.125emX}}

\usepackage[T1]{fontenc}
\usepackage[utf8]{inputenc}
\usepackage{float,enumitem,tcolorbox,booktabs,multirow,tabularx,array,tikz,listings}
\usepackage{xcolor}
\usepackage{microtype}
\usetikzlibrary{arrows.meta,positioning,fit,backgrounds}
\tcbuselibrary{listings,skins,breakable}
\newcommand{\mbertid}{\texttt{bert-\allowbreak base-\allowbreak multi\allowbreak lingual-\allowbreak cased}}
\newcommand{\pdtbcsv}{\texttt{english\_\allowbreak discourse\_\allowbreak pdtb3.csv}}
\usepackage{url}
\makeatletter
\g@addto@macro\UrlBreaks{\do\a\do\b\do\c\do\d\do\e\do\f\do\g\do\h\do\i\do\j\do\k\do\l\do\m\do\n\do\o\do\p\do\q\do\r\do\s\do\t\do\u\do\v\do\w\do\x\do\y\do\z\do\0\do\1\do\2\do\3\do\4\do\5\do\6\do\7\do\8\do\9}
\makeatother
\usepackage[hidelinks,bookmarks=false]{hyperref}
\newcommand{\rel}[3]{%
\noindent\textbf{#1}\\
\noindent\textit{Arg1:} #2\\
\noindent\textit{Arg2:} #3\\[3pt]}
\let\year\Accessyear
\year{}
\vol{}
\renewcommand{\thetable}{\Roman{table}}
\begin{document}
\history{}
\doi{}
\title{Mitigating Data Scarcity for Low-Resource Discourse Classification via Large Language Model-Driven Cross-Lingual Synthetic Generation}
\author{\uppercase{Muhammad Owais Mahmood}\authorrefmark{1},
\uppercase{Syed Ali Ahmed}\authorrefmark{1},
\uppercase{Hassaan Muhammad Khan}\authorrefmark{1},
\uppercase{Bahram Khan Baloch}\authorrefmark{1,2}, AND
\uppercase{Muhammad Ibrahim Chhipa}\authorrefmark{1}}
\address[1]{ByteCorp, Karachi, Pakistan (e-mail: owais.mahmood@bytecorp.io; syed.ali@bytecorp.io; hassaan.khan@bytecorp.io; ibrahim.chippa@bytecorp.io)}
\address[2]{Universit\"at des Saarlandes, Saarbr\"ucken, Germany}
\markboth{Mahmood \headeretal: LLM-Driven Cross-Lingual Synthetic Generation}{Mahmood \headeretal: LLM-Driven Cross-Lingual Synthetic Generation}
\corresp{Corresponding author: Muhammad Owais Mahmood (e-mail: owais.mahmood@bytecorp.io).}
\begin{abstract}
Urdu discourse-relation classification is limited by the lack of annotated training data.
This study examines whether synthetic examples generated by large language models can support the task when Urdu annotations are unavailable.
Examples from the English Penn Discourse Treebank 3.0 provide few-shot anchors for generating labeled English and Urdu argument pairs.
We fine-tune five multilingual encoders on individual and pooled synthetic datasets and evaluate them on held-out English data and a human-verified Urdu test set.
On Urdu, pooled synthetic training achieves 44.15\% accuracy with the best encoder, exceeding the 28.77\% majority-class baseline and direct English-to-Urdu transfer for all five encoders.
In size-matched experiments, human-supervised Urdu training remains stronger, reaching 68.24\% accuracy with the best encoder.
The larger synthetic dataset yields modest gains but does not close this gap.
These results indicate that synthetic supervision is useful when Urdu annotations are unavailable, while human annotation remains more effective when it can be obtained.
\end{abstract}

\titlepgskip=-21pt
\maketitle

\section{Introduction}

\begin{figure*}[t]
    \centering
    \includegraphics[width=5.3in]{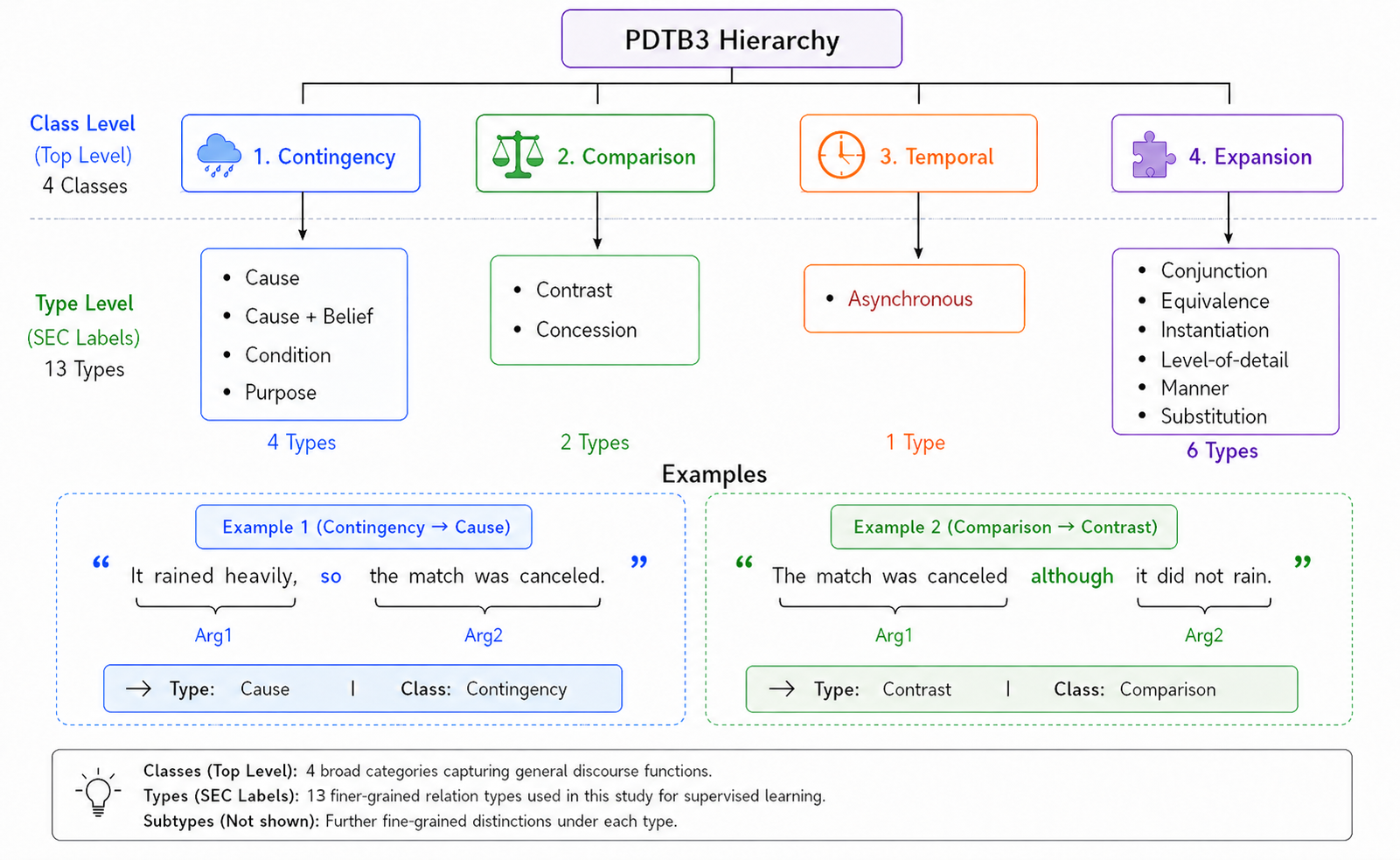}
    \caption{Hierarchy of the four top-level and 13 second-level discourse relations in Penn Discourse Treebank 3.0 (PDTB3).}
    \label{fig:pdtb3_hierarchy}
\end{figure*}

Natural language processing (NLP) resources remain concentrated in a small set of high-resource languages~\cite{pakray2025nlp}.
Urdu has over 200 million speakers, yet its annotated datasets remain scarce or fragmented.
Its right-to-left script, verb-final word order, and morphological complexity complicate direct adaptation from English~\cite{8636191,khan2019urdu}.

Existing Urdu resources address part-of-speech tagging~\cite{8636191,khan2019urdu}, sentiment classification~\cite{ahmed2022urdu,khan2022multi}, and machine translation~\cite{israr2024improved,ahmed2024urdu}.
Higher-level linguistic structure has received less attention.
Recent work covers Urdu Discourse Representation Structures~\cite{amin-etal-2025-evaluating, amin-etal-2025-semantic}.
However, Penn Discourse Treebank (PDTB)-style classification between two argument spans remains unaddressed.
This task requires annotators to distinguish fine-grained logical relations.
Such expertise makes corpus construction expensive, especially where annotation infrastructure is limited.

The PDTB~\cite{prasad-etal-2008-penn} and its third edition (PDTB3)~\cite{webber-etal-2019-pdtb3} provide a widely used framework for this task, organizing relations into a hierarchical structure. At the top level, relations are grouped into four broad classes (\textit{Contingency}, \textit{Comparison}, \textit{Temporal}, and \textit{Expansion}). These classes are then refined into more specific types, as illustrated in Fig.~\ref{fig:pdtb3_hierarchy}. For example, in \textit{``It rained heavily, so the match was canceled,''} the relation is \textit{causal}, which belongs to the \textit{Contingency} class, while in \textit{``The match was canceled although it did not rain,''} the relation is \textit{contrastive} and belongs to the \textit{Comparison} class. \textit{Temporal} relations capture ordering such as before or after, and \textit{Expansion} includes relations such as elaboration or addition. In some cases, these are further divided into subtypes that capture finer distinctions such as directionality or argument roles.

In this study, we target the \textit{Type} level of the PDTB3 hierarchy, hereafter referred to as second-level (SEC) labels, consisting of 13 relation types. This level provides a good balance for supervised learning: it is more informative than the four top-level classes, while avoiding the lack of sufficient training examples associated with finer-grained subtypes. Applications of discourse classification include text summarization, question answering, argument mining, and machine translation~\cite{see-etal-2017-get,rajpurkar-etal-2016-squad}. However, the absence of any large-scale discourse-annotated corpus for Urdu means that these applications remain largely underexplored for the language.

Under our SEC-level scheme, each example consists of two argument spans, \textit{Arg1} and \textit{Arg2}, with a relation label from the SEC taxonomy; worked examples for all four top-level families are provided in the appendix (Fig.~\ref{fig:pdtb3_examples}). For instance, in the \textsc{Contingency.Cause} example, \textit{Arg2} (``the entire street was flooded'') expresses the consequence of the situation described in \textit{Arg1} (``the pipe burst after the temperature dropped below freezing'').

English often marks discourse relations with connectives such as \textit{because}, \textit{however}, and \textit{meanwhile}.
Urdu can express the same relations through morphology, syntax, and context.
These differences limit direct transfer from English surface patterns.

We study whether large language model (LLM)-generated supervision can help discourse relation classification when target-language annotations are unavailable.
GPT-4o, Llama~3.3 70B, DeepSeek-R1, and Mistral-Large generate PDTB3-labeled argument pairs in English and Urdu.
We train multilingual encoders on individual generators and on a pooled synthetic training set.
We evaluate every English condition on the PDTB3 test partition.
We evaluate every Urdu condition on the same human-verified Urdu test set.
The Urdu benchmark is a resource contribution alongside the generation pipeline: it provides a common real-text evaluation of discourse classification across training sources and encoders.

Our analysis separates supervision quality from scale.
The size-matched ablation compares equal quantities of synthetic and human training data.
The full-scale condition then measures the benefit of the larger pooled synthetic training set.
This design tests whether synthetic scale offsets lower per-example utility.

Our key contributions are as follows:

\begin{itemize}
    \item We introduce a PDTB3-guided pipeline for generating English and Urdu discourse-relation training data with four LLMs.

    \item We separate per-example utility from scale through size-matched and full-scale synthetic training conditions.

    \item We construct a human-verified Urdu discourse-relation benchmark from news and encyclopedic text, providing a shared test set of 2,763 instances across 13 selected SEC relations for evaluating synthetic and human supervision.

    \item We evaluate synthetic supervision on real English and human-verified Urdu test sets across five multilingual encoders.

    \item We quantify where synthetic scale helps and where human supervision retains an advantage.
\end{itemize}

\section{Related Work}

Large annotated datasets and pretrained transformers have advanced many NLP tasks~\cite{NIPS2017_3f5ee243,devlin-etal-2019-bert,10010807}.
Applications include question answering~\cite{rajpurkar-etal-2016-squad,chen-etal-2017-reading} and summarization~\cite{see-etal-2017-get,lewis-etal-2020-bart}.
They also include translation~\cite{NIPS2017_3f5ee243,wu2016googlesneuralmachinetranslation}, sentiment analysis~\cite{socher-etal-2013-recursive,tang-etal-2016-aspect,albladi2026twssentinovelhybridframework}, and classification~\cite{devlin-etal-2019-bert,yang-etal-2016-hierarchical}.
Most resources remain concentrated in high-resource languages~\cite{pakray2025nlp}.
Urdu work covers tagging~\cite{8636191,khan2019urdu}, sentiment analysis~\cite{khan2022multi,ahmed2022urdu}, and translation~\cite{israr2024improved,ahmed2024urdu}.

Discourse classification assigns coherence relations between text spans~\cite{mann1988rhetorical}.
Relations can express causality, temporal order, contrast, or elaboration.
Rhetorical Structure Theory represents document-level hierarchical structures~\cite{mann1988rhetorical}.
PDTB annotates pairwise argument relations anchored by connectives or adjacency~\cite{prasad-etal-2008-penn,prasad-etal-2014-reflections}.
PDTB3 supports comparable resources in Chinese, Hindi, Arabic, Czech, and Turkish~\cite{webber-etal-2019-pdtb3,prasad-etal-2014-reflections}.
The Conference on Computational Natural Language Learning (CoNLL) shared tasks standardized English and Chinese evaluation~\cite{xue-etal-2015-conll, xue-etal-2016-conll}.
No equivalent Urdu benchmark is available.

Recent work applies LLMs to discourse classification.
Maekawa et al.~\cite{maekawa-etal-2024-obtain} fine-tuned Llama~2 with quantized low-rank adaptation (QLoRA) for Rhetorical Structure Theory (RST) classification.
Li et al.~\cite{li-etal-2024-dialogue} formulated dialogue discourse classification as sequence generation with T0.
Pastor et al.~\cite{pastor-etal-2025-enhancing} used LLM-generated data for English social-media RST parsing.
Our study instead examines Urdu, where no discourse-relation training corpus is available.

LLMs increasingly generate synthetic data for low-resource NLP~\cite{pakray2025nlp}.
Synthetic text can be lexically restricted and distributionally smoother than real writing.
Classifiers may therefore learn generator-specific shortcuts.
Reasoning-oriented models such as DeepSeek-R1 may produce longer and more varied outputs~\cite{deepseekai2025r1}.
Translate-train offers another strategy by translating an annotated high-resource corpus.
mBART-50 supports translation across 50 languages, including South Asian languages~\cite{liu2020mbart}.

Our downstream models include \mbertid{}~\cite{devlin-etal-2019-bert}, which covers 104 languages.
Multilingual Bidirectional Encoder Representations from Transformers (mBERT)'s shared subword representations support cross-lingual transfer~\cite{pires2019multilingual}.
We also evaluate MuRIL (Multilingual Representations for Indian Languages), mDeBERTa-v3 (multilingual decoding-enhanced BERT with disentangled attention), Glot500 (a multilingual model covering over 500 languages), and XLM-R (XLM-RoBERTa)~\cite{conneau-etal-2020-unsupervised}.
This comparison tests whether conclusions depend on one encoder.

\section{Methodology}
\label{sec:metho}

\begin{figure*}[t]
    \centering
    \begin{tikzpicture}[
        box/.style={draw, rounded corners, align=center, text width=4.1cm, minimum height=1.3cm, font=\small},
        flow/.style={-{Latex}, thick}, node distance=0.65cm and 0.7cm]
    \node[box] (anchors) {English PDTB3 anchors\\Three examples per relation};
    \node[box, right=of anchors] (generation) {Four LLM generators\\13 selected SEC relations};
    \node[box, right=of generation] (synthetic) {Urdu synthetic training\\6,421 size-matched\\69,395 full-scale};
    \node[box, below=of anchors] (sources) {Real Urdu text\\Wikipedia, Jang, Dawn};
    \node[box, below=of generation] (review) {LLM segmentation and labelling\\Agreement filtering\\Human review and adjudication};
    \node[box, below=of synthetic] (real) {Evaluated Urdu split\\6,421 training; 2,763 test\\2,758 distinct test pairs};
    \node[box, below=of real] (evaluation) {Five multilingual encoders\\13 classifier outputs\\Shared Urdu test set};
    \draw[flow] (anchors) -- (generation);
    \draw[flow] (generation) -- (synthetic);
    \draw[flow] (sources) -- (review);
    \draw[flow] (review) -- (real);
    \draw[flow] (real) -- (evaluation);
    \draw[flow] (synthetic.east) -- ++(0.4,0) |- (evaluation.east);
    \end{tikzpicture}
    \caption{End-to-end pipeline. Three English PDTB3 examples per relation guide generation. Four LLMs each produce $\sim$18K labeled pairs per language. Wikipedia, Jang, and Dawn provide real Urdu text. Inter-model agreement filtering and human review construct the Urdu benchmark. The evaluated Urdu files contain 6,421 training and 2,763 test instances. Classifiers encode each pair as \texttt{[CLS] Arg1 [SEP] Arg2 [SEP]}. Generation and classification use the same 13 selected SEC relations. English and Urdu conditions use their respective real held-out test sets.}
    \label{fig:pipeline}
\end{figure*}
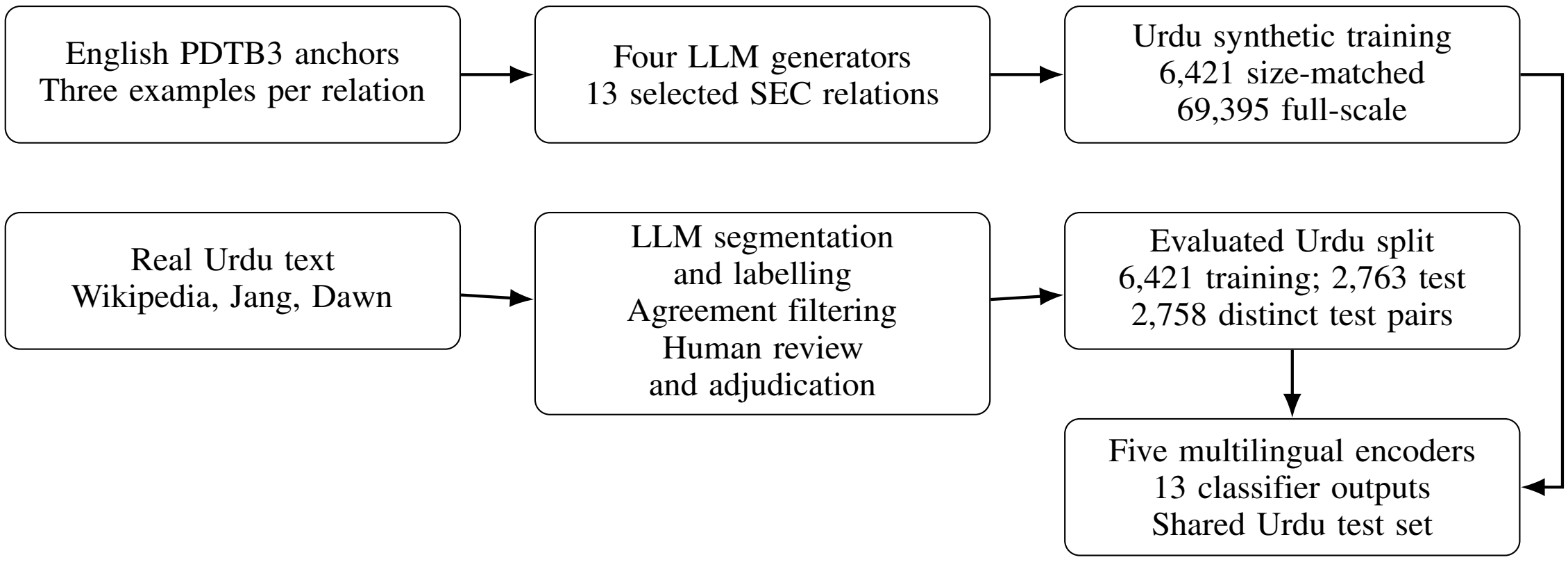

We target 13 second-level discourse relations during generation.
The evaluated English and Urdu classifiers use the same 13 selected SEC labels.
No prior Urdu corpus, test set, or task baseline is available.
We adopt the Penn Discourse Treebank 3.0 annotation scheme~\cite{prasad-etal-2008-penn}.
PDTB3 contains 53,631 Wall Street Journal relation tokens~\cite{webber-etal-2019-pdtb3}.
GPT-4o, Llama~3.3, Mistral-Large, and DeepSeek-R1 generate labeled pairs for the 13 selected SEC relations.
Each generator produces $\sim$18{,}000 pairs per language.
We use these pairs to fine-tune multilingual classifiers.
Fig.~\ref{fig:pipeline} summarizes the pipeline.

\subsection{LLM Selection: Benchmarking Synthetic Data Quality}
\label{sec:selection}

We compare four LLMs under one generation and downstream evaluation protocol.
The selection covers proprietary and open-weight model families.
GPT-4o~\cite{openai2023gpt4} and DeepSeek-R1~\cite{deepseekai2025r1} provide reasoning-oriented generators.
Llama~3.3~\cite{dubey2024llama3} and Mistral-Large~\cite{jiang2023mistral} provide open-weight alternatives.

DeepSeek-R1 uses large-scale reinforcement learning without intermediate supervised fine-tuning~\cite{deepseekai2025r1}.
Reasoning-oriented training can produce longer, more structured outputs~\cite{wei2023chainofthoughtpromptingelicitsreasoning,marjanovic2026thoughtology}.
Such outputs may help generate coherent argument pairs.
We treat this explanation as a hypothesis and compare all four generators empirically.

\subsection{Synthetic Data Generation}
\label{sec:valid}

We generate data for 13 second-level PDTB3 labels, which we call SEC labels.
The four top-level classes merge relations such as \textsc{Contingency.Purpose} and \textsc{Contingency.Condition}.
Third-level subtypes provide finer detail but contain fewer examples~\cite{lin2009recognizing}.
The SEC level balances semantic detail with trainable class support.

\subsubsection{The Few-Shot Anchor Strategy}
The generator must express the requested logical relation, not merely fluent text. We provide three randomly sampled English PDTB3 examples per relation as in-context demonstrations~\cite{brown2020fewshot, dong2022survey}.
These examples act as structural anchors for the requested relation.

The method assumes that a discourse relation retains its logical structure across languages.
For example, \textsc{Contingency.Cause} links a cause with its consequence.
English anchors specify this schema while the prompt requests target-language surface forms.
Fig.~\ref{lst:anchorfig} provides the template.

The prompt imposes semantic, structural, and linguistic constraints.
The closed label set and anchors constrain the requested relation.
The format \texttt{SEC -- Arg1 -- Arg2} supports deterministic parsing.
Language and register instructions limit surface-form drift.

\subsubsection{English as a Validation Control}
\label{sec:english-control}
We first apply the generation procedure to English.
The PDTB3 standard partition provides a real held-out test set for synthetic-to-real evaluation.
We train the supervised reference on 80\% of the real annotated English data.
English performance tests the generation procedure but does not predict Urdu performance.
We therefore analyze the two languages separately.

Each LLM generated approximately 18,000 English argument pairs.
An 80/20 split produced 14,400 training and 3,600 validation instances per generator.
Restricting PDTB3 to the 13 selected SEC relations yields about 18,000 annotated pairs.
Each individual synthetic source therefore approximates the real-corpus scale.
The pooled synthetic training set is larger because it combines all four sources.
Table~\ref{tab:encoders} reports accuracy on the PDTB3 test partition.

\subsection{Urdu Human-Verified Test Set Construction}
\label{sec:gold}

Urdu lacks an existing discourse-relation benchmark.
We therefore construct a real-text evaluation set from Urdu Wikipedia, Jang, and Dawn.
These sources cover formal encyclopedic and news registers.

An LLM segmented the scraped text into candidate \textit{Arg1} and \textit{Arg2} pairs.
GPT-4o and DeepSeek-R1 independently assigned SEC labels.
Pairs receiving the same label from both models were retained as silver-labeled candidates for human review.
This filter selects agreement but does not establish independent label correctness.

Two Urdu annotators reviewed the retained relation labels and argument boundaries.
A senior linguist adjudicated disagreements, and unresolved cases were excluded.
The resulting human-verified corpus supplies all real-text Urdu evaluation in this paper.
The evaluated human-supervised training set contains 6,421 usable instances, and the shared test set contains 2,763 instances.
These counts refer to the examples used in the reported experiments, rather than intermediate collection or annotation files.
The test set contains 2,758 distinct argument pairs.

\subsection{Urdu Label Distributions}
\label{sec:urdu-distributions}

Table~\ref{tab:urdu-label-counts} presents the available counts for the 13 relations targeted during generation, separately for the human training and test sets, each generator, and the pooled set.
The table describes the selected relations in the available dataset summaries; it does not establish the classifier output inventory.
The table reports class counts for the human training and test sets, each generator, and the pooled set, without percentages or totals.

\begin{table*}[t]
\centering
\scriptsize
\setlength{\tabcolsep}{3pt}
\renewcommand{\arraystretch}{1.15}
\caption{Urdu counts for the 13 relations targeted during generation. Human train/test denote the human-verified datasets.}
\label{tab:urdu-label-counts}
\begin{tabular}{@{}lrrrrrrr@{}}
\toprule
Relation & Human train & Human test & GPT-4o & Llama 3.3 & Mistral-Large & DeepSeek-R1 & Pooled \\
\midrule
Comparison.Concession    & 426   & 47  & 1,448 & 1,323 & 1,378 & 1,200 & 4,838 \\
Comparison.Contrast      & 934   & 397 & 983   & 2,215 & 1,606 & 715   & 5,182 \\
Contingency.Cause        & 1,248 & 533 & 4,242 & 2,643 & 1,839 & 4,403 & 12,554 \\
Contingency.Cause+Belief & 70    & 31  & 566   & 613   & 757   & 152   & 1,648 \\
Contingency.Condition    & 211   & 84  & 236   & 1,272 & 1,341 & 142   & 2,602 \\
Contingency.Purpose      & 162   & 67  & 1,133 & 1,131 & 1,384 & 1,125 & 4,441 \\
Expansion.Conjunction   & 1,831 & 795 & 3,484 & 2,237 & 1,747 & 3,728 & 10,648 \\
Expansion.Equivalence   & 61    & 28  & 306   & 1,215 & 988   & 264   & 2,659 \\
Expansion.Instantiation & 491   & 84  & 1,192 & 1,322 & 1,325 & 1,196 & 4,888 \\
Expansion.Level-of-detail & 1,022 & 468 & 2,539 & 0   & 1,545 & 2,663 & 6,525 \\
Expansion.Manner        & 7     & 0   & 666   & 855   & 1,422 & 635   & 3,242 \\
Expansion.Substitution  & 39    & 9   & 388   & 494   & 1,203 & 355   & 2,330 \\
Temporal.Asynchronous   & 472   & 203 & 1,486 & 3,698 & 1,459 & 996   & 7,020 \\
\bottomrule
\end{tabular}
\end{table*}

Within the test-set summary, \textsc{Expansion.Conjunction}, \textsc{Contingency.Cause}, and \textsc{Expansion.Level-of-detail} are the most frequent displayed relations.
\textsc{Expansion.Manner} has no test examples, so this test set cannot measure performance on that relation.
The summaries also report no \textsc{Expansion.Level-of-detail} examples for Llama~3.3, despite its presence in the test set and the other generator summaries.
These coverage differences motivate reporting macro-F1 alongside accuracy, but label counts alone do not establish relation-specific classification performance.

\subsection{Downstream Classifier: mBERT Fine-Tuning}
\label{sec:mbert}

We fine-tune \mbertid{}~\cite{devlin-etal-2019-bert} with a linear classification head.
The input format is \texttt{[CLS] Arg1 [SEP] Arg2 [SEP]}.
The classifier receives the \texttt{[CLS]} representation.
Within each encoder comparison, all conditions use the same training configuration.

We use mBERT because it covers English and Urdu within one multilingual encoder~\cite{pires2019multilingual}.
Its local inference also avoids recurring application programming interface (API) calls.
We repeat the data-source comparisons with MuRIL, mDeBERTa-v3, Glot500, and XLM-R.
We scope conclusions to the encoders for which the reported separation holds.

\section{Results and Analysis}

\subsection{Experimental Setup and Evaluation Metrics}

English uses the 13 selected PDTB3 SEC labels.
Urdu uses the same 13 selected PDTB3 SEC labels.
Training conditions include each generator, the pooled synthetic training set, and human-supervised references.
Every English condition uses the same real PDTB3 test partition.
Every Urdu condition uses the same 2,763-instance human-verified test set.
We report accuracy and macro-F1 because the label distribution is imbalanced.
The Urdu size-matched table reports accuracy and macro-F1. Appendix~\ref{app:urdu-perclass} provides Urdu XLM-R per-class results across three seeds; Appendix~\ref{app:perclass} documents the English mDeBERTa-v3 per-class results and their evaluation limitations.
Table~\ref{tab:encoders} contains accuracy.
The Urdu majority-class accuracy is 0.2877.

\begin{table*}[t]
\centering
\small
\caption{Classification accuracy across five encoders on real held-out test sets. English models use the PDTB3 standard test partition. Urdu models use the 2,763-instance human-verified test set (Section~\ref{sec:gold}), whose majority-class accuracy is 0.2877. Each English synthetic source uses 14,400 training instances. The full-scale pooled Urdu training set contains 69,395 usable instances; the size-matched comparison uses 6,421 training instances per condition. \textsuperscript{\dag}Human-annotated English PDTB3 training. \textsuperscript{\ddag}Human-verified Urdu training. EN denotes English and UR denotes Urdu. Bold marks the best encoder within each displayed source row.}
\label{tab:encoders}
\setlength{\tabcolsep}{3pt}
\renewcommand{\arraystretch}{1.2}
\begin{tabular}{@{}ll cc cc cc cc cc@{}}
\toprule
& & \multicolumn{2}{c}{\textbf{mBERT}}
  & \multicolumn{2}{c}{\textbf{MuRIL}}
  & \multicolumn{2}{c}{\textbf{mDeBERTa-v3}}
  & \multicolumn{2}{c}{\textbf{Glot500}}
  & \multicolumn{2}{c}{\textbf{XLM-R}} \\
\cmidrule(lr){3-4}\cmidrule(lr){5-6}\cmidrule(lr){7-8}\cmidrule(lr){9-10}\cmidrule(lr){11-12}
\textbf{Source} & \textbf{Generator}
  & EN & UR & EN & UR & EN & UR & EN & UR & EN & UR \\
\midrule
Real\textsuperscript{\dag}  & Human (PDTB3)   & 0.5127 & --     & 0.3974 & --     & \textbf{0.5503} & --     & 0.4975 & --     & 0.5408 & --     \\
Human\textsuperscript{\ddag} & Human verified & --     & 0.6417 & --     & 0.4148 & --     & \textbf{0.6728} & --     & 0.6402 & --     & 0.6638 \\
\midrule
\multirow{4}{*}{Synthetic}
  & GPT-4o        & 0.4416 & 0.3865 & 0.4746 & 0.3572 & 0.5492 & 0.3981 & 0.5117 & 0.3699 & \textbf{0.5764} & \textbf{0.4072} \\
  & Llama~3.3     & 0.1531 & 0.3272 & 0.1597 & 0.4101 & \textbf{0.4260} & \textbf{0.4260} & 0.1556 & 0.3768 & 0.1783 & 0.3927 \\
  & Mistral-Large & 0.3930 & 0.3418 & 0.4448 & 0.3764 & 0.4894 & 0.3590 & 0.4514 & 0.3663 & \textbf{0.5132} & \textbf{0.4021} \\
  & DeepSeek-R1   & 0.5085 & 0.3557 & 0.5127 & 0.3446 & \textbf{0.7333} & 0.3771 & 0.6838 & 0.3637 & 0.7305 & \textbf{0.3931} \\
\midrule
Pooled & All LLMs & 0.5202 & 0.3972 & 0.7089 & 0.3941 & 0.7496 & 0.4191 & 0.7117 & 0.4169 & \textbf{0.7606} & \textbf{0.4415} \\
\bottomrule
\end{tabular}
\end{table*}

\subsection{English Validation Under Full-Scale Synthetic Training}

Table~\ref{tab:encoders} reports accuracy for the full-scale training conditions.
The pooled synthetic condition contains more training examples than either human reference.
Its result therefore measures the practical full-data setting, not equal-size supervision quality.

\paragraph{DeepSeek-R1 under mBERT.}
Under mBERT, DeepSeek-R1-generated English training reaches 0.5085 accuracy on the PDTB3 test set.
The human-trained mBERT reference reaches 0.5127.
This difference is 0.0042 accuracy (0.42 percentage points).
The comparison is specific to English, mBERT, and the displayed training sizes.
It does not establish equal per-example utility.

Under mBERT, Llama~3.3 reaches 0.1531, Mistral-Large reaches 0.3930, and GPT-4o reaches 0.4416.
Llama~3.3 and Mistral-Large both exceed 0.90 on their synthetic validation sets.

\paragraph{Encoder-specific behavior.}
We repeat the comparison with MuRIL, mDeBERTa-v3, Glot500, and XLM-R.
On English, DeepSeek-R1 reaches 0.7333 with mDeBERTa-v3, compared with 0.5085 under mBERT.
The pooled English condition uses more training data than the human reference, so it is not an equal-size comparison.

The Urdu pattern differs.
The pooled full-scale condition reaches 0.3972, 0.3941, 0.4191, 0.4169, and 0.4415 under mBERT, MuRIL, mDeBERTa-v3, Glot500, and XLM-R.
The controlled human references reach 0.6417, 0.4148, 0.6728, 0.6402, and 0.6638.
An earlier mBERT human value of 0.4011 did not reproduce under the controlled pipeline.
Across three controlled seeds, mBERT human training instead reaches $0.6411 \pm 0.0027$.
MuRIL remains an unstable exception, while the four reliable encoders retain a large human-supervision advantage.

\paragraph{Linguistic statistics and transfer.}
Table~\ref{tab:linguistic} documents the linguistic statistics of each
generator's output.
DeepSeek-R1 has a mean combined length of 166.14 characters and a vocabulary of 33,846 English tokens.
Real PDTB3 has a mean length of 181.27 characters and a vocabulary of 43,101 tokens.
Llama~3.3 and Mistral-Large produce shorter arguments and smaller vocabularies.
These descriptive similarities coincide with DeepSeek-R1's English ranking but do not establish causality.

\begin{table}[t]
\centering
\caption{Descriptive statistics for synthetic English data and real PDTB3. Avg.\ Len is mean combined argument length in characters. Vocab is the number of unique tokens. Readability is Flesch Reading Ease, where lower values indicate greater complexity. These statistics do not control for training-set size or establish a causal relationship with transfer performance.}
\label{tab:linguistic}
\setlength{\tabcolsep}{5pt}
\renewcommand{\arraystretch}{1.25}
\begin{tabular}{@{}lccc@{}}
\toprule
\textbf{Source} & \textbf{Avg.\ Len} & \textbf{Vocab} & \textbf{Readability} \\
\midrule
Real (PDTB3)         & 181.27 & 43,101 & 47.45 \\
\textbf{DeepSeek-R1} & \textbf{166.14} & \textbf{33,846} & \textbf{47.57} \\
GPT-4o               & 113.60 & 23,076 & 49.99 \\
Mistral-Large        & 63.75  & 15,405 & 79.22 \\
Llama~3.3            & 79.94  & 8,731  & 62.42 \\
\bottomrule
\end{tabular}
\end{table}

\subsection{Applying the Pipeline to Urdu}
\label{sec:urdu}

We apply the same generation procedure to Urdu.
Every Urdu condition uses the same 2,763-instance human-verified test set.
The majority-class accuracy is 0.2877.
The displayed synthetic conditions exceed this baseline.

Urdu differs from English in word order, pro-drop, morphology, modality, and evidentiality.
mBERT also contains less Urdu than English pre-training data~\cite{pires2019multilingual}.
The real Urdu test set contains formal news and Wikipedia text.

\paragraph{Generator rankings depend on the encoder.}
GPT-4o is the strongest individual Urdu generator under mBERT and XLM-R, reaching 0.3865 and 0.4072 accuracy, respectively.
Llama~3.3 instead leads under MuRIL, mDeBERTa-v3, and Glot500, reaching 0.4101, 0.4260, and 0.3768.
DeepSeek-R1 does not lead any of the Urdu encoder comparisons in Table~\ref{tab:encoders}.
Thus, the Urdu results do not identify one universally strongest generator: generator choice interacts with the downstream encoder.
These are descriptive rankings, not statistically established differences.

\paragraph{Pooling helps in some encoder settings.}
The full-scale pooled set exceeds the strongest individual generator under mBERT, Glot500, and XLM-R by 0.0107, 0.0401, and 0.0343 accuracy, respectively.
For MuRIL and mDeBERTa-v3, Llama~3.3 remains numerically ahead of pooling by 0.0160 and 0.0069.
Pooling therefore does not uniformly improve Urdu classification.
Because the pooled condition combines generators and uses a different training-set size, this comparison cannot isolate the benefit of generator diversity.

\paragraph{Cross-lingual transfer without Urdu data.}
We apply the English human-trained classifiers to the Urdu test set without Urdu fine-tuning (Table~\ref{tab:transfer}; per-class results are in Appendix~\ref{app:urdu-perclass}).
Accuracy is 0.3594 for mBERT, 0.2555 for MuRIL, 0.3659 for mDeBERTa-v3, 0.3464 for Glot500, and 0.4285 for XLM-R.
All five scores remain below the full-scale pooled synthetic results, with gaps ranging from 0.0130 for XLM-R to 0.1386 for MuRIL.
Zero-shot transfer is also lower than the size-matched pooled synthetic mean for four encoders. XLM-R is the exception, but its transfer accuracy of 0.4285 is close to the pooled mean of $0.4198 \pm 0.0245$.
MuRIL falls below the 0.2877 majority-class baseline. Its predictions cover only four of the 14 labels, while XLM-R obtains zero F1 for several relations, including \textsc{Contingency.Condition}.
These results show that Urdu synthetic training is generally more effective than direct English-to-Urdu transfer; human-supervised Urdu models remain substantially stronger.

\begin{table*}[t]
\centering
\small
\setlength{\tabcolsep}{8pt}
\caption{Zero-shot cross-lingual transfer: English human-trained (PDTB3) classifiers evaluated on the 2,763-instance Urdu test set without further training, compared with Urdu pooled synthetic and Urdu human training (6,421 examples each, mean over three seeds). Labels used is the number of the 14 output labels the transferred model predicts at least once. Transfer macro-F1 averages over 14 labels for mBERT and XLM-R and over 13 for the other three encoders, so it is not strictly comparable across rows. Majority-class accuracy is 0.2877 (macro-F1 0.0319).}
\label{tab:transfer}
\begin{tabular}{@{}lcccrr@{}}
\toprule
& \multicolumn{3}{c}{EN human $\rightarrow$ UR transfer} & \multicolumn{2}{c}{UR accuracy} \\
\cmidrule(lr){2-4}\cmidrule(lr){5-6}
Encoder & Accuracy & Macro-F1 & Labels used & Pooled synth. & Human \\
\midrule
mBERT       & 0.3594 & 0.1678 & 12 & 0.3767 & 0.6411 \\
MuRIL       & 0.2555 & 0.0706 & 4  & 0.3263 & 0.4125 \\
mDeBERTa-v3 & 0.3659 & 0.1602 & 10 & 0.3918 & 0.6720 \\
Glot500     & 0.3464 & 0.1269 & 8  & 0.3883 & 0.6273 \\
XLM-R       & 0.4285 & 0.1908 & 12 & 0.4198 & 0.6824 \\
\bottomrule
\end{tabular}
\end{table*}

\paragraph{Accuracy and coverage across relations.}
The size-matched results in Table~\ref{tab:size-matched} provide a complementary view through macro-F1, which gives equal weight to each label included in the metric.
For pooled synthetic training, XLM-R has the highest mean accuracy and macro-F1 among the evaluated encoders, whereas mBERT has the highest human-supervised macro-F1 despite XLM-R having the highest human-supervised accuracy.
Under mBERT, mDeBERTa-v3, Glot500, and XLM-R, human supervision exceeds pooled synthetic supervision on both metrics.
MuRIL shows a different pattern: human training has higher accuracy, but its mean macro-F1 is slightly lower, with a difference small relative to the reported seed variability.
These aggregate metrics do not identify which Urdu relations account for the differences; that requires the per-class reporting discussed below.

The full-scale pooled synthetic set contains 69,395 usable training instances.
It reaches 0.3972, 0.3941, 0.4191, 0.4169, and 0.4415 under mBERT, MuRIL, mDeBERTa-v3, Glot500, and XLM-R, respectively.
These results use the five-epoch training protocol.
They measure the practical full-data setting rather than equal-size supervision quality.

\subsection{Size-Matched Ablation: Per-Example Utility}
\label{sec:size-matched}

The full pooled condition measures performance when synthetic generation supplies more training data.
It does not isolate per-example utility.
We therefore compare human and pooled synthetic training with exactly 6,421 usable rows per condition.
The sampler preserves label proportions and uses the same test set, optimizer, and five-epoch training configuration.
We repeat the comparison across five encoders with seeds 42, 1337, and 2024.
Standard deviations describe variability across seeds; they do not establish statistical equivalence or significance.
Table~\ref{tab:size-matched} reports mean and standard deviation on the 2,763-instance Urdu test set.
The evaluated pipeline uses 13 labels, and the majority-class accuracy is 0.2877.

\begin{table*}[t]
\centering
\small
\caption{Size-matched Urdu ablation on the 2,763-instance human-verified test set. Pooled synthetic and human training each use exactly 6,421 examples. Values are mean $\pm$ standard deviation across seeds 42, 1337, and 2024. The evaluated pipeline uses 13 labels, and the majority-class accuracy is 0.2877. Bold marks the higher mean within each encoder and metric; differences are descriptive unless accompanied by a named statistical test.}
\label{tab:size-matched}
\setlength{\tabcolsep}{5pt}
\begin{tabular}{lcccc}
\toprule
& \multicolumn{2}{c}{Accuracy} & \multicolumn{2}{c}{Macro-F1} \\
\cmidrule(lr){2-3}\cmidrule(lr){4-5}
Encoder & Pooled synthetic & Human & Pooled synthetic & Human \\
\midrule
mBERT & $0.3767 \pm 0.0055$ & $\mathbf{0.6411 \pm 0.0027}$ & $0.2344 \pm 0.0111$ & $\mathbf{0.5391 \pm 0.0014}$ \\
MuRIL & $0.3263 \pm 0.0270$ & $\mathbf{0.4125 \pm 0.0335}$ & $\mathbf{0.1285 \pm 0.0214}$ & $0.1216 \pm 0.0315$ \\
mDeBERTa-v3 & $0.3918 \pm 0.0120$ & $\mathbf{0.6720 \pm 0.0014}$ & $0.2296 \pm 0.0079$ & $\mathbf{0.5056 \pm 0.0350}$ \\
Glot500 & $0.3883 \pm 0.0051$ & $\mathbf{0.6273 \pm 0.0136}$ & $0.1931 \pm 0.0213$ & $\mathbf{0.3159 \pm 0.0303}$ \\
XLM-R & $0.4198 \pm 0.0245$ & $\mathbf{0.6824 \pm 0.0170}$ & $0.2569 \pm 0.0195$ & $\mathbf{0.5120 \pm 0.0783}$ \\
\bottomrule
\end{tabular}
\end{table*}

For mBERT, mDeBERTa-v3, Glot500, and XLM-R, equal-sized human training exceeds pooled synthetic training by 0.2390--0.2801 accuracy.
The corresponding macro-F1 advantage is 0.1228--0.3047.
MuRIL is an exception: both conditions remain weak and have overlapping variability in macro-F1.
We therefore exclude MuRIL from general claims about the synthetic--human gap.

The size-matched results establish lower per-example utility for pooled synthetic supervision under the four reliable encoders.
The full-scale scores exceed the corresponding size-matched synthetic means by 0.0205--0.0286 accuracy for those encoders.
This comparison suggests rather than establishes a scale benefit.
Moreover, the larger dataset produces more optimizer updates within the same five-epoch protocol.
The result therefore measures the practical benefit of full-data training, not a compute-matched causal effect of data diversity.
Even at full scale, human supervision retains a 0.2104--0.2529 accuracy advantage under the four reliable encoders.

\subsection{Per-Class Analysis}
Table~\ref{tab:urdu-perclass} reports Urdu precision, recall, and F1 for XLM-R, averaged across three seeds with 6,421 training examples per condition.
Human supervision has higher mean F1 on frequent relations including \textsc{Expansion.Conjunction}, \textsc{Contingency.Cause}, and \textsc{Expansion.Level-of-detail}.
Pooled synthetic supervision has higher mean F1 on \textsc{Expansion.Instantiation}, \textsc{Contingency.Cause+Belief}, and \textsc{Expansion.Equivalence}, although performance remains low on these relations.
Thus, the aggregate human-supervision advantage does not extend uniformly to every relation.
These comparisons are descriptive means and do not establish statistical significance.
Detailed Urdu results appear in Appendix~\ref{app:urdu-perclass}; the English per-class results appear in Appendix~\ref{app:perclass}.

\section{Conclusion and Future Work}

We presented an LLM-based pipeline for generating English and Urdu discourse-relation training data.
Four generators use English PDTB3 examples as structural anchors~\cite{openai2023gpt4,dubey2024llama3,deepseekai2025r1,jiang2023mistral,webber-etal-2019-pdtb3}.
Multilingual encoders then learn from individual or pooled synthetic training sets.

The evidence supports a bounded conclusion.
The three-seed size-matched analysis shows that pooled synthetic supervision is weaker per example under four reliable encoders.
Full-scale synthetic training improves accuracy by 0.0205--0.0286 over the corresponding size-matched means in this comparison.
These gains narrow but do not remove the human-supervision advantage.
Because full-scale training also uses more optimizer updates, the results suggest a practical scale benefit rather than isolate a causal data-volume effect.
Generator rankings also differ between English and Urdu.
Synthetic validation performance therefore does not determine real-text transfer quality.

Synthetic supervision remains useful when target-language annotation is unavailable.
It does not replace human annotation when comparable human training data exists.
Future work should test whether these patterns extend to other low-resource languages.

\section*{Limitations}
The 2,763-instance Urdu test set covers formal news and Wikipedia text.
It does not measure conversational, dialectal, or code-switched Urdu.
A broader test collection would determine whether the results transfer across registers.

The test-set pipeline begins with LLM segmentation and agreement-based silver labels.
This construction can favor examples that the two labelling models already find easy.
Independent blind annotation would measure and reduce that selection effect.
Records needed to calculate inter-annotator agreement, label-correction rates, exclusion counts, and annotation quality before human review are unavailable.
Consequently, we cannot quantify annotator consistency or the extent to which human review changed the initial labels.
Human verification describes the reported review procedure, rather than a measured level of annotation reliability.

Full-scale differences can reflect optimization variance, and comparisons against size-matched means remain descriptive.
The same five-epoch protocol also gives the larger synthetic dataset more optimizer updates; a compute-matched comparison is needed to isolate the effect of additional distinct examples.

The evaluation covers the 13 selected SEC labels rather than the full PDTB3 inventory.
The Urdu test set contains 2,763 instances but only 2,758 distinct argument pairs; the reported scores are computed over instances.

The Urdu per-class analysis covers XLM-R only. Reports for the other encoders and reconciliation of macro-F1 label-averaging conventions are needed before generalizing these relation-level findings.

\section*{Ethical Considerations}
Urdu annotators reviewed labels and argument boundaries, with senior adjudication of disagreements (Section~\ref{sec:gold}).
The source text comes from Urdu Wikipedia, Jang, and Dawn.
Detailed collection records, including article-level source URLs, collection dates, and scraping and preprocessing logs, are unavailable, limiting reproducibility of the corpus construction.
Documentation establishing permission or applicable reuse terms for the collected newspaper material is also unavailable; this paper therefore makes no claim that such permission was obtained.
Any release of argument pairs or source text is conditional on establishing the applicable redistribution rights.
It will not redistribute source material beyond the applicable licenses.
Appendix~\ref{app:e2e} provides the prompts and one worked example.
We plan to release the permitted datasets and the generation and training code after publication.

\setcounter{dbltopnumber}{3}\renewcommand{\dbltopfraction}{0.9}
\appendices
\section{Implementation Details of the Synthetic Generation Pipeline}
\label{app:e2e}

\providecommand{\gloss}[1]{\textit{#1}}
\lstdefinestyle{promptbox}{
  basicstyle=\ttfamily\footnotesize,
  breaklines=true,
  columns=fullflexible,
  frame=single,
  framesep=4pt,
  xleftmargin=6pt,
  xrightmargin=6pt,
  showstringspaces=false,
  keepspaces=true,
  captionpos=b
}

\begin{figure}[t]
\centering
\begin{tcolorbox}[
    colback=gray!6,
    colframe=gray!50,
    boxrule=0.6pt,
    arc=3pt,
    title={\textbf{PDTB3 Discourse Relation Examples}},
    fonttitle=\small\bfseries
]
\small
\rel{Contingency.Cause}
    {The pipe burst after the temperature dropped below freezing.}
    {The entire street was flooded within minutes.}
\rel{Comparison.Concession}
    {She trained every day for six months.}
    {She still did not qualify for the finals.}
\rel{Temporal.Asynchronous}
    {The committee reviewed the proposal over the weekend.}
    {A decision was announced on Monday morning.}
\rel{Expansion.Instantiation}
    {The policy affects all public sector employees.}
    {This includes teachers, nurses, and civil administrators.}
\rel{Contingency.Condition}
    {You must submit the application before the deadline.}
    {Otherwise your file will not be considered.}
\rel{Comparison.Contrast}
    {The report was thorough and well-documented.}
    {The board rejected it without explanation.}
\rel{Expansion.Conjunction}
    {The flood damaged roads across the northern districts.}
    {Several bridges were declared unsafe for use.}
\rel{Contingency.Purpose}
    {The team worked through the night.}
    {They wanted to meet the morning deadline.}
\end{tcolorbox}
\caption{PDTB3 discourse relation examples showing Arg1 and Arg2 spans with their Class.Type SEC-level labels.}
\label{fig:pdtb3_examples}
\end{figure}

This appendix traces one relation from source examples to classifier output.
It records each prompt, intermediate artifact, and transformation used by the implementation.
The following templates describe the prompts for the 13 targeted relations.
The generation pipeline targets \textbf{13 second-level (SEC) relations}.
The classifier uses \texttt{[CLS] Arg1 [SEP] Arg2 [SEP]}.
The evaluated Urdu label mapping contains the same 13 selected SEC relations, yielding 13 classifier outputs.

\subsection{The Pipeline at a Glance}
\label{app:e2e:overview}

The pipeline has five stages: anchors, generation, Urdu benchmark construction, encoder fine-tuning, and evaluation.
The first two stages create synthetic training data.
The third stage constructs the real-text Urdu benchmark.
The final stages train and evaluate the classifier.

\subsection{Stage-by-Stage: What Goes In, What Comes Out}
\label{app:e2e:stages}

\paragraph{\textcircled{1} Few-Shot Anchors (input $\rightarrow$ structural anchors).}
\emph{Input:} the English PDTB-3 seed corpus
(\pdtbcsv{}, 17{,}781 rows;
columns \texttt{Top, Sec, Conn, Arg1, Arg2}).
\emph{Operation:} for the target relation, sample three annotated English argument
pairs of the same relation (class-filtered,
\texttt{random\_state=42}).
\emph{Output:} a JSON array of three
\texttt{\{Sec, Arg1, Arg2\}} objects injected into the prompt as
\texttt{\{examples\}}. These anchors specify the relational structure rather than
the surface wording and are reused for both English and Urdu generation.

\paragraph{\textcircled{2} Synthetic Data Generation (anchors $\rightarrow$ pairs).}
\emph{Input:} the rendered prompt (anchors, target
\texttt{\{language\}}, and target
\texttt{\{sec\_value\}}).
\emph{Models:} GPT-4o, Llama~3.3 (70B), Mistral-Large, and DeepSeek-R1,
each generating independently.
\emph{Output contract:} one example per line in the format
\texttt{Sec - Arg1 - Arg2}, with both arguments written in the target language.
\emph{Quality gate:} outputs are retained only if they contain exactly three
fields, the relation label is valid, and it matches the requested target.
\emph{Result:} approximately 18{,}000 English examples per generator, split 80/20 into training and validation. For Urdu, the full-scale pooled training condition uses 69,395 usable instances, while the size-matched pooled condition uses 6,421 (Table~\ref{tab:size-matched}).

\paragraph{\textcircled{3} Urdu Test Set (real text $\rightarrow$ human-verified set).}
\emph{Input:} real Urdu text scraped from Urdu Wikipedia, Jang, and Dawn.
\emph{Extraction:} an LLM segments candidate \texttt{Arg1}/\texttt{Arg2} pairs.
\emph{Silver labelling by agreement:} GPT-4o and DeepSeek label each pair independently.
Pairs with matching model-assigned labels are retained as silver-labeled candidates for human review.
Two Urdu annotators review each retained relation and segmentation.
A senior linguist adjudicates disagreements, and unresolved cases are removed.
The evaluated files contain \textbf{6{,}421} usable human-supervised training instances and \textbf{2{,}763} human-verified test instances, including 2,758 distinct test argument pairs (Section~\ref{sec:gold}). These are experimental dataset sizes, not counts of intermediate annotation candidates.

\paragraph{\textcircled{4} mBERT Fine-Tuning (pairs $\rightarrow$ classifier).}
Each argument pair is encoded as
\texttt{[CLS] Arg1 [SEP] Arg2 [SEP]} and processed by
\mbertid{} with a linear classification head over the
13 SEC labels.

\paragraph{\textcircled{5} Evaluation (classifier $\rightarrow$ scores).}
English models use the PDTB-3 test partition.
Urdu models use the same human-verified Urdu test set.
We report accuracy, macro-averaged precision, recall, and F1.
Human-trained conditions provide supervised references rather than absolute upper bounds.

\subsection{Prompt Templates}
\label{app:e2e:prompts}

\paragraph{All-relations generation prompt.}
Six system messages; used by the GPT-4o/DeepSeek/Mistral generators (Fig.~\ref{lst:allrel}).

\paragraph{Structural-anchor generation prompt (Fig.~\ref{lst:anchorfig}).}
Conditions on a single target relation \texttt{\{sec\_value\}} and injects three
class-filtered anchor pairs as \texttt{\{examples\}} (see also Fig.~\ref{lst:anchor}).

\paragraph{Urdu-set silver-labelling prompt.}
Temperature 0, batches of five; used in stage (3) to label real scraped pairs
(run once per model, kept only on agreement; Fig.~\ref{lst:label}).

\begin{figure}[H]
\centering
\begin{tcolorbox}[
    colback=gray!6,
    colframe=gray!50,
    boxrule=0.6pt,
    arc=3pt,
    top=2pt, bottom=2pt, left=4pt, right=4pt,
    title={\textbf{All-relations generation prompt}},
    fonttitle=\small\bfseries
]
\ttfamily\footnotesize
\setlength{\parskip}{4pt}
\setlength{\parindent}{0pt}

\textbf{[system]} You are creating a discourse parsing dataset for \{language\}. Please follow the structure and instructions carefully.

\textbf{[system]} Given the following examples from an English dataset: \{examples\}

\textbf{[system]} These are the classes for `sec' values (connection types):
\begin{itemize}[leftmargin=4mm, topsep=0pt, itemsep=4pt, parsep=0pt, partopsep=0pt, label={-}]
\item Temporal.Asynchronous
\item Contingency.Cause
\item Comparison.Contrast
\item Contingency.Cause+Belief
\item Contingency.Condition
\item Contingency.Purpose
\item Comparison.Concession
\item Expansion.Conjunction
\item Expansion.Equivalence
\item Expansion.Instantiation
\item Expansion.Level-of-detail
\item Expansion.Manner
\item Expansion.Substitution
\end{itemize}
\textbf{[system]} Make sure the `sec' values generated are from the above list, and the following text corresponds to the sec value.

\textbf{[system]} For each example, output only the following values in the format: sec - Arg1 - Arg2. Do not include any numbering or additional commentary.

\textbf{[system]} Ensure Arg1 and Arg2 are strictly in \{language\}. Return only the values without explanations or extra text.
\end{tcolorbox}
\caption{All-relations generation prompt}
\label{lst:allrel}
\end{figure}

\begin{figure}[H]
\centering
\begin{tcolorbox}[
    colback=gray!6,
    colframe=gray!50,
    boxrule=0.6pt,
    arc=3pt,
    title={\textbf{Structural-anchor per-relation prompt}},
    fonttitle=\small\bfseries
]
\ttfamily\footnotesize
\noindent
[system] You are creating a discourse parsing dataset in \{language\}.\\[4pt]
[system] Given these examples from an English dataset (JSON array):\\
\hspace*{9mm}\{examples\}\\[4pt]
[system] Valid Sec classes: - \{valid\_secs\_gen\}\\[4pt]
[system] Generate exactly one example per line, in this format: Sec - Arg1 - Arg2.\\[4pt]
[system] Arg1 and Arg2 must be in \{language\}.\\[4pt]
[system] Do NOT include numbering, bullets, or blank lines. Return only the lines.
\end{tcolorbox}
\caption{Structural-anchor per-relation prompt}
\label{lst:anchor}
\end{figure}

\begin{figure}[H]
\centering
\begin{tcolorbox}[
    colback=gray!6,
    colframe=gray!50,
    boxrule=0.6pt,
    arc=3pt,
    top=2pt, bottom=2pt, left=4pt, right=4pt,
    title={\textbf{Structural-anchor prompt}},
    fonttitle=\small\bfseries
]
\ttfamily\footnotesize
\setlength{\parskip}{4pt}
\setlength{\parindent}{0pt}

\textbf{System:} You are a linguist specialising in discourse annotation. You are constructing a PDTB3 dataset for \{language\}.

\textbf{Structural anchors:} The following English PDTB3 examples illustrate the target relations: \{examples\}

\textbf{Instruction:} Using the relational schemas demonstrated above, generate novel sentence pairs in \{language\} satisfying the following discourse relation: \{sec\_value\}

\textbf{Valid SEC classes:} Temporal.Asynchronous, Contingency.Cause, Contingency.Cause+Belief, Contingency.Condition, Contingency.Purpose, Comparison.Contrast, Comparison.Concession, Expansion.Conjunction, Expansion.Equivalence, Expansion.Instantiation, Expansion.Level-of-detail, Expansion.Manner, Expansion.Substitution (13 classes)

\textbf{Constraints:} (i) Output strictly in \{language\}; maintain formal journalistic or encyclopedic register. (ii) Arg1 and Arg2 must satisfy the logical constraints of the specified relation. (iii) Format: sec - Arg1 - Arg2. No preamble or commentary.
\end{tcolorbox}
\caption{Structural-anchor prompt}
\label{lst:anchorfig}
\end{figure}

\begin{figure}[H]
\centering
\begin{tcolorbox}[
    colback=gray!6,
    colframe=gray!50,
    boxrule=0.6pt,
    arc=3pt,
    title={\textbf{Silver-labelling prompt }},
    fonttitle=\small\bfseries
]
\ttfamily\footnotesize
\noindent
[system] You are a discourse relation classifier. Classify each Arg1 and Arg2\\
\hspace*{9mm}pair into one of the following Sec values:\\
\hspace*{9mm}- Temporal.Asynchronous\\
\hspace*{9mm}- Contingency.Cause\\
\hspace*{9mm}... (full SEC inventory) ...\\
\hspace*{9mm}- Expansion.Substitution\\[4pt]
[user]\hspace*{3mm}Here are the examples (one per line) in the format:\\
\hspace*{9mm}``Arg1'',``Arg2''\\
\hspace*{9mm}\{examples\}\\[4pt]
\hspace*{9mm}Only output the results as CSV lines in the format: ``Arg1'',``Arg2'',Sec.\\
\hspace*{9mm}Use double quotes around Arg1 and Arg2 to handle commas.\\
\hspace*{9mm}Do not include any headers or extra text.
\end{tcolorbox}
\caption{Silver-labelling prompt}
\label{lst:label}
\end{figure}
\subsection{A Complete Worked Example (\textsc{Contingency.Cause}, Urdu)}
\label{app:e2e:worked}

We trace \textsc{Contingency.Cause} from seed corpus to final prediction.
The anchors come from \pdtbcsv{}.
The output is one generated Urdu row from the planned release.

\paragraph{Step 1: Sample structural anchors.}
Three \textsc{Contingency.Cause} pairs are drawn from the English seed and
serialized into \texttt{\{examples\}} (Fig.~\ref{lst:fewshot}):

\begin{figure}[H]
\centering

\begin{tcblisting}{
    colback=gray!6,
    colframe=gray!50,
    boxrule=0.6pt,
    arc=3pt,
    listing only,
    title={\textbf{Few-shot anchor examples (JSON format)}},
    fonttitle=\small\bfseries,
    listing options={
        basicstyle=\ttfamily\footnotesize,
        breaklines=true,
        breakatwhitespace=true,
        columns=fullflexible,
        keepspaces=true,
        showstringspaces=false
    }
}
[
  {"Sec":"Contingency.Cause",
   "Arg1":"Mr. Asman should make distinctions among Fidel, the army and the Cuban people",
   "Arg2":"They are not interchangeable"},

  {"Sec":"Contingency.Cause",
   "Arg1":"They are not interchangeable, since they are motivated to act based on their own circumstances",
   "Arg2":"It is naivete to equate a government's policies with the will of the people"},

  {"Sec":"Contingency.Cause",
   "Arg1":"In terms of sheer brutality, the Somali regime of Siad Barre may rank as No. 1 in the world",
   "Arg2":"The only reason that Somalia remains in obscurity is numbers"}
]
\end{tcblisting}

\caption{Few-shot anchor examples passed as \texttt{\{examples\}} in the generation prompt (JSON format).}
\label{lst:fewshot}

\end{figure}

\paragraph{Step 2: Render the prompt.}
With \texttt{\{language\}=\allowbreak Urdu} and \texttt{\{sec\_value\}=\allowbreak Contingency.\allowbreak Cause}, together with the anchors above, the instantiated prompt is Fig.~\ref{lst:anchorfig} with those values substituted.

\paragraph{Step 3: Raw model output.}

The LLM returns dash-delimited lines. The first (shown here with an English
gloss standing in for the Urdu original) is:
\begin{quote}\ttfamily\footnotesize
Contingency.Cause - \gloss{He worked hard} - \gloss{He passed the exam}
\end{quote}
The generated pair mirrors the anchors' \emph{logical (causal) form}, not their
vocabulary.

\paragraph{Step 4: Parse and filter.}

The generated line satisfies the validation criteria (correct format and target
relation) and is therefore retained; malformed or off-target outputs are discarded.

\paragraph{Step 5: Appended dataset row.}

The kept pair is written to the per-LLM CSV (schema \texttt{Sec, Arg1, Arg2}):
\begin{quote}\ttfamily\footnotesize
Contingency.Cause, \gloss{He worked hard}, \gloss{He passed the exam}
\end{quote}
This is one of the $\approx$18{,}000 pairs for this generator/language.

\paragraph{Step 6: mBERT encoding.}

\begin{quote}\ttfamily\footnotesize
{[}CLS{]} \gloss{He worked hard} {[}SEP{]} \gloss{He passed the exam} {[}SEP{]}
\end{quote}
The \texttt{[CLS]} vector feeds the linear classification head.
The evaluated Urdu mapping contains 13 outputs, one for each selected SEC relation.

\begin{table*}[t]
\centering
\small
\setlength{\tabcolsep}{6pt}
\caption{Urdu XLM-R per-class results: mean precision (P), recall (R), and F1 across three seeds, with 6,421 training examples per condition. N is test support.}
\label{tab:urdu-perclass}
\begin{tabular}{@{}lr rrr rrr@{}}
\toprule
& & \multicolumn{3}{c}{Human} & \multicolumn{3}{c}{Pooled synthetic} \\
\cmidrule(lr){3-5}\cmidrule(lr){6-8}
Relation & N & P & R & F1 & P & R & F1 \\
\midrule
Expansion.Conjunction & 795 & 0.777 & 0.717 & 0.746 & 0.533 & 0.566 & 0.548 \\
Contingency.Cause & 533 & 0.672 & 0.770 & 0.717 & 0.480 & 0.533 & 0.503 \\
Expansion.Level-of-detail & 468 & 0.606 & 0.683 & 0.642 & 0.389 & 0.294 & 0.334 \\
Comparison.Contrast & 397 & 0.642 & 0.715 & 0.676 & 0.465 & 0.192 & 0.271 \\
Temporal.Asynchronous & 203 & 0.655 & 0.764 & 0.705 & 0.406 & 0.647 & 0.499 \\
Contingency.Condition & 84 & 0.722 & 0.790 & 0.754 & 0.594 & 0.099 & 0.167 \\
Expansion.Instantiation & 84 & 0.222 & 0.008 & 0.015 & 0.127 & 0.254 & 0.169 \\
Contingency.Purpose & 67 & 0.625 & 0.736 & 0.675 & 0.658 & 0.169 & 0.265 \\
Comparison.Concession & 47 & 0.639 & 0.305 & 0.397 & 0.163 & 0.553 & 0.251 \\
Contingency.Cause+Belief & 31 & 0.000 & 0.000 & 0.000 & 0.288 & 0.108 & 0.157 \\
Expansion.Equivalence & 28 & 0.000 & 0.000 & 0.000 & 0.127 & 0.167 & 0.143 \\
Expansion.Substitution & 9 & 0.830 & 0.556 & 0.663 & 0.186 & 0.667 & 0.290 \\
\bottomrule
\end{tabular}
\end{table*}

\begin{table*}[t]
\centering
\small
\setlength{\tabcolsep}{4pt}
\caption{Urdu per-class F1 for all five encoders, mean over seeds 42, 1337, and 2024, with 6,421 training examples per condition. H denotes human-verified Urdu training and P denotes pooled synthetic training. N is test support.}
\label{tab:urdu-f1-all}
\begin{tabular}{@{}lr cc cc cc cc cc@{}}
\toprule
& & \multicolumn{2}{c}{mBERT} & \multicolumn{2}{c}{MuRIL} & \multicolumn{2}{c}{mDeBERTa-v3} & \multicolumn{2}{c}{Glot500} & \multicolumn{2}{c}{XLM-R} \\
\cmidrule(lr){3-4}\cmidrule(lr){5-6}\cmidrule(lr){7-8}\cmidrule(lr){9-10}\cmidrule(lr){11-12}
Relation & N & H & P & H & P & H & P & H & P & H & P \\
\midrule
Expansion.Conjunction & 795 & 0.706 & 0.501 & 0.654 & 0.409 & 0.735 & 0.534 & 0.717 & 0.529 & 0.746 & 0.548 \\
Contingency.Cause & 533 & 0.677 & 0.426 & 0.478 & 0.426 & 0.720 & 0.423 & 0.679 & 0.469 & 0.717 & 0.503 \\
Expansion.Level-of-detail & 468 & 0.608 & 0.305 & 0.162 & 0.000 & 0.617 & 0.263 & 0.581 & 0.064 & 0.642 & 0.334 \\
Comparison.Contrast & 397 & 0.599 & 0.240 & 0.039 & 0.258 & 0.685 & 0.295 & 0.618 & 0.135 & 0.676 & 0.271 \\
Temporal.Asynchronous & 203 & 0.651 & 0.415 & 0.003 & 0.171 & 0.685 & 0.399 & 0.652 & 0.453 & 0.704 & 0.499 \\
Contingency.Condition & 84 & 0.696 & 0.127 & 0.276 & 0.132 & 0.702 & 0.136 & 0.659 & 0.159 & 0.754 & 0.167 \\
Expansion.Instantiation & 84 & 0.231 & 0.159 & 0.000 & 0.000 & 0.050 & 0.163 & 0.000 & 0.069 & 0.015 & 0.169 \\
Contingency.Purpose & 67 & 0.598 & 0.248 & 0.000 & 0.323 & 0.558 & 0.200 & 0.200 & 0.318 & 0.675 & 0.265 \\
Comparison.Concession & 47 & 0.468 & 0.278 & 0.000 & 0.000 & 0.380 & 0.188 & 0.000 & 0.284 & 0.397 & 0.251 \\
Contingency.Cause+Belief & 31 & 0.000 & 0.120 & 0.000 & 0.000 & 0.000 & 0.141 & 0.000 & 0.021 & 0.000 & 0.157 \\
Expansion.Equivalence & 28 & 0.000 & 0.180 & 0.000 & 0.036 & 0.000 & 0.187 & 0.000 & 0.049 & 0.000 & 0.143 \\
Expansion.Substitution & 9 & 0.775 & 0.283 & 0.045 & 0.000 & 0.473 & 0.284 & 0.000 & 0.154 & 0.663 & 0.290 \\
\bottomrule
\end{tabular}
\end{table*}

\begin{table*}[t]
\centering
\small
\setlength{\tabcolsep}{8pt}
\caption{Per-class F1 of the zero-shot English$\rightarrow$Urdu transfer classifiers (English human-trained, evaluated on the Urdu test set) for all five encoders. N is test support.}
\label{tab:transfer-perclass}
\begin{tabular}{@{}lrccccc@{}}
\toprule
Relation & N & mBERT & MuRIL & mDeBERTa-v3 & Glot500 & XLM-R \\
\midrule
Expansion.Conjunction & 795 & 0.499 & 0.402 & 0.532 & 0.522 & 0.595 \\
Contingency.Cause & 533 & 0.380 & 0.388 & 0.378 & 0.355 & 0.495 \\
Expansion.Level-of-detail & 468 & 0.321 & 0.000 & 0.283 & 0.252 & 0.312 \\
Comparison.Contrast & 397 & 0.206 & 0.000 & 0.078 & 0.030 & 0.199 \\
Temporal.Asynchronous & 203 & 0.348 & 0.045 & 0.354 & 0.269 & 0.404 \\
Contingency.Condition & 84 & 0.000 & 0.000 & 0.000 & 0.000 & 0.000 \\
Expansion.Instantiation & 84 & 0.163 & 0.000 & 0.060 & 0.130 & 0.267 \\
Contingency.Purpose & 67 & 0.058 & 0.083 & 0.056 & 0.000 & 0.058 \\
Comparison.Concession & 47 & 0.121 & 0.000 & 0.096 & 0.093 & 0.138 \\
Contingency.Cause+Belief & 31 & 0.000 & 0.000 & 0.000 & 0.000 & 0.000 \\
Expansion.Equivalence & 28 & 0.125 & 0.000 & 0.000 & 0.000 & 0.000 \\
Expansion.Substitution & 9 & 0.127 & 0.000 & 0.245 & 0.000 & 0.203 \\
\bottomrule
\end{tabular}
\end{table*}

\begin{table*}[t]
\centering
\small
\setlength{\tabcolsep}{6pt}
\caption{English mDeBERTa-v3 per-class precision (P), recall (R), and F1. N is test support.}
\label{tab:perlabel}
\begin{tabular}{@{}lr rrr rrr@{}}
\toprule
& & \multicolumn{3}{c}{DeepSeek-R1} & \multicolumn{3}{c}{Pooled synthetic} \\
\cmidrule(lr){3-5}\cmidrule(lr){6-8}
Relation & N & P & R & F1 & P & R & F1 \\
\midrule
Contingency.Cause & 1,350 & 0.763 & 0.781 & 0.772 & 0.763 & 0.827 & 0.794 \\
Expansion.Conjunction & 1,050 & 0.753 & 0.812 & 0.781 & 0.799 & 0.816 & 0.807 \\
Expansion.Level-of-detail & 814 & 0.726 & 0.722 & 0.724 & 0.773 & 0.720 & 0.746 \\
Expansion.Instantiation & 348 & 0.761 & 0.761 & 0.761 & 0.738 & 0.761 & 0.750 \\
Comparison.Concession & 340 & 0.652 & 0.721 & 0.684 & 0.725 & 0.688 & 0.706 \\
Contingency.Purpose & 332 & 0.715 & 0.885 & 0.791 & 0.729 & 0.876 & 0.796 \\
Temporal.Asynchronous & 321 & 0.770 & 0.763 & 0.767 & 0.773 & 0.813 & 0.792 \\
Comparison.Contrast & 213 & 0.706 & 0.709 & 0.712 & 0.697 & 0.737 & 0.715 \\
Expansion.Substitution & 103 & 0.650 & 0.631 & 0.640 & 0.645 & 0.689 & 0.667 \\
Expansion.Equivalence & 70 & 0.540 & 0.286 & 0.374 & 0.435 & 0.386 & 0.409 \\
Contingency.Cause+Belief & 57 & 0.615 & 0.140 & 0.229 & 0.300 & 0.053 & 0.090 \\
Contingency.Condition & 41 & 0.789 & 0.732 & 0.759 & 0.751 & 0.726 & 0.738 \\
\bottomrule
\end{tabular}
\end{table*}

\begin{table*}[t]
\centering
\small
\setlength{\tabcolsep}{7pt}
\caption{English per-class F1 for MuRIL, mDeBERTa-v3, and Glot500. D denotes DeepSeek-R1 synthetic training and P denotes pooled synthetic training. N is test support.}
\label{tab:eng-f1-all}
\begin{tabular}{@{}lr cc cc cc@{}}
\toprule
& & \multicolumn{2}{c}{MuRIL} & \multicolumn{2}{c}{mDeBERTa-v3} & \multicolumn{2}{c}{Glot500} \\
\cmidrule(lr){3-4}\cmidrule(lr){5-6}\cmidrule(lr){7-8}
Relation & N & D & P & D & P & D & P \\
\midrule
Contingency.Cause & 1,350 & 0.671 & 0.782 & 0.772 & 0.794 & 0.738 & 0.769 \\
Expansion.Conjunction & 1,050 & 0.601 & 0.775 & 0.781 & 0.807 & 0.748 & 0.766 \\
Expansion.Level-of-detail & 814 & 0.254 & 0.662 & 0.724 & 0.746 & 0.635 & 0.668 \\
Expansion.Instantiation & 348 & 0.291 & 0.701 & 0.761 & 0.750 & 0.689 & 0.711 \\
Comparison.Concession & 340 & 0.327 & 0.652 & 0.684 & 0.706 & 0.591 & 0.616 \\
Contingency.Purpose & 332 & 0.743 & 0.787 & 0.791 & 0.796 & 0.793 & 0.806 \\
Temporal.Asynchronous & 321 & 0.613 & 0.728 & 0.767 & 0.792 & 0.736 & 0.751 \\
Comparison.Contrast & 213 & 0.194 & 0.651 & 0.712 & 0.715 & 0.599 & 0.632 \\
Expansion.Substitution & 103 & 0.034 & 0.645 & 0.640 & 0.667 & 0.539 & 0.648 \\
Expansion.Equivalence & 70 & 0.000 & 0.305 & 0.374 & 0.409 & 0.244 & 0.312 \\
Contingency.Cause+Belief & 57 & 0.000 & 0.105 & 0.229 & 0.090 & 0.033 & 0.145 \\
Contingency.Condition & 41 & 0.402 & 0.667 & 0.759 & 0.738 & 0.706 & 0.765 \\
\bottomrule
\end{tabular}
\end{table*}

\paragraph{Step 7: Prediction and scoring.}
The predicted relation is compared with the human-adjudicated label and contributes
to the reported accuracy and macro-averaged precision, recall, and F1.

\section{Urdu Per-Class Results}
\label{app:urdu-perclass}

Table~\ref{tab:urdu-perclass} gives the full per-class results for the Urdu XLM-R experiments, and Table~\ref{tab:urdu-f1-all} reports per-class F1 for all five encoders.
We report per-class results for the encoder with the highest human-supervised accuracy in each language: XLM-R for Urdu and mDeBERTa-v3 for English (Table~\ref{tab:encoders} and Table~\ref{tab:size-matched}).

Table~\ref{tab:transfer-perclass} gives per-class F1 for the zero-shot transfer classifiers of all five encoders.
Transfer scores are highest on the frequent relations \textsc{Expansion.Conjunction}, \textsc{Contingency.Cause}, and \textsc{Temporal.Asynchronous}, and are zero for \textsc{Contingency.Condition} and \textsc{Contingency.Cause+Belief} under every encoder.

\section{English Per-Class Results}
\label{app:perclass}

Table~\ref{tab:perlabel} reports mDeBERTa-v3 results, comparing DeepSeek-R1 training with pooled synthetic training.
mDeBERTa-v3 has the highest English human-supervised accuracy among the five encoders (Table~\ref{tab:encoders}). Table~\ref{tab:eng-f1-all} reports per-class F1 for MuRIL, mDeBERTa-v3, and Glot500.

\twocolumn[\relax]
\raggedbottom
\makeatletter
\let\oldthebibliography\thebibliography
\renewcommand\thebibliography[1]{\oldthebibliography{#1}\setlength{\parsep}{0pt}\setlength{\topsep}{0pt}\setlength{\partopsep}{0pt}\setlength{\itemsep}{0pt}\setlength{\parskip}{0pt}}
\makeatother
\bibliographystyle{IEEEtran}
\bibliography{references}

\begin{IEEEbiography}[{\includegraphics[width=1in,height=1.25in,clip,keepaspectratio]{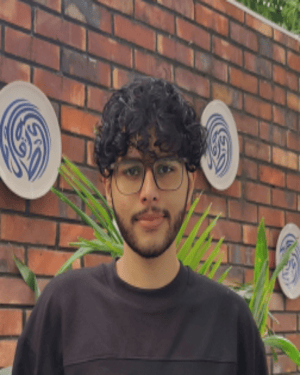}}]{Muhammad Owais Mahmood}
received the B.S. degree in software engineering from DHA Suffa University, Karachi, Pakistan, in 2025, where the University Gold Medal was awarded. M. O. Mahmood is currently pursuing the M.Sc. degree in Cognitive Systems: Language, Learning and Reasoning at the University of Potsdam, Germany. Since 2024, M. O. Mahmood has been with ByteCorp, Karachi, first as a Research Engineer and, since April 2025, as a Machine Learning Engineer, working on edge-deployed perception and multimodal conversational systems. Research interests include computer vision, natural language processing, machine translation, multimodal learning, efficient inference on edge devices, and knowledge distillation.
\end{IEEEbiography}

\begin{IEEEbiography}[{\includegraphics[width=1in,height=1.25in,clip,keepaspectratio]{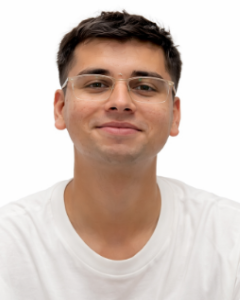}}]{Syed Ali Ahmed}
received the bachelor's degree in artificial intelligence from the National University of Computer and Emerging Sciences (FAST-NUCES), Pakistan. Research interests include machine learning, computer vision, machine unlearning, anomaly detection, and energy-based models, as well as natural language processing and synthetic data generation for low-resource languages. S. A. Ahmed has coauthored work on fraudulent job-posting detection and on cross-domain generalization in machine unlearning, the latter presented at the ECCV 2026 Workshop on Unlearning and Model Editing.
\end{IEEEbiography}

\begin{IEEEbiography}[{\includegraphics[width=1in,height=1.25in,clip,keepaspectratio]{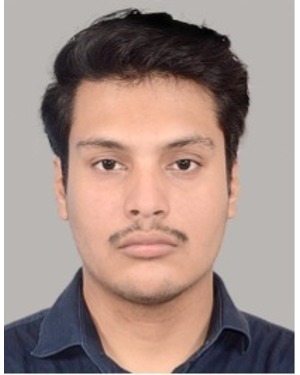}}]{Hassaan Muhammad Khan}
received the B.E. degree in electrical engineering (artificial intelligence) from the National University of Sciences and Technology, Pakistan, in 2025, where the Chancellor's Silver Medal was awarded. H. M. Khan is a Machine Learning Engineer at ByteCorp, Karachi, Pakistan, and was previously a Visiting Researcher with the VILA Lab at the Mohamed bin Zayed University of Artificial Intelligence. Published work includes papers at ACL 2025 and SCAP 2024. Research interests include multimodal artificial intelligence, vision--language models, retrieval-augmented generation, and natural language processing.
\end{IEEEbiography}

\begin{IEEEbiography}[{\includegraphics[width=1in,height=1.25in,clip,keepaspectratio]{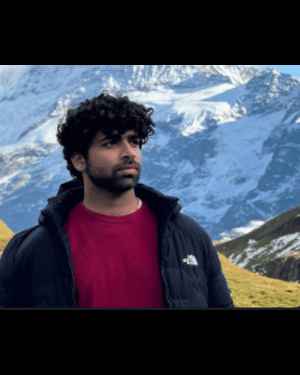}}]{Bahram Khan Baloch}
is a graduate student in computer science at Saarland University, Germany. Research interests include machine learning, deep learning, computer vision, and large language models. B. K. Baloch has coauthored work on cost-sensitive learning for imbalanced classification, weight de-correlation as regularization in deep networks, spatial-economic analysis of electric-vehicle charging-station placement, and Arabic--English machine translation.
\end{IEEEbiography}

\begin{IEEEbiography}[{\includegraphics[width=1in,height=1.25in,clip,keepaspectratio]{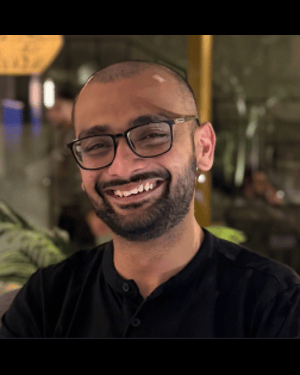}}]{Muhammad Ibrahim Chhipa}
received the bachelor's degree in computer science from the National University of Computer and Emerging Sciences (FAST-NUCES), Pakistan. M. I. Chhipa is a Co-Founder and the Chief Technology Officer of ByteCorp. Research interests include Arabic--English machine translation and the integration of multilingual sequence-to-sequence models with large language models. M. I. Chhipa coauthored ``ALEF: An Enhanced Approach to Arabic-English Bilingual Translation,'' presented at the 2025 10th International Conference on Machine Learning Technologies (ICMLT), Helsinki, Finland.
\end{IEEEbiography}

\EOD
\end{document}